\documentclass{article} % For LaTeX2e
\usepackage{iclr2023_conference_tinypaper,times}

\usepackage{amsmath,amsfonts,bm}

\def\eqref#1{equation~\ref{#1}}
\def\1{\bm{1}}

\DeclareMathAlphabet{\mathsfit}{\encodingdefault}{\sfdefault}{m}{sl}
\SetMathAlphabet{\mathsfit}{bold}{\encodingdefault}{\sfdefault}{bx}{n}

\usepackage{hyperref}
\usepackage{url}
\usepackage{algorithm}
\usepackage{algpseudocode}
\usepackage{color,soul}

\title{MetaXLR - Mixed Language Meta Representation Transformation for Low-resource Cross-lingual Learning based on Multi-Armed Bandit}

\author{Liat Bezalel, Eyal Orgad\thanks{Both authors contributed equally} \\
Department of Computer Science\\
Tel-Aviv University, Israel\\
\texttt{\{liatbezalel, eyalorgad\}@mail.tau.ac.il}
}

\iclrfinalcopy 
\begin{document}

\maketitle

\begin{abstract}

Transfer learning for extremely low-resource languages is a challenging task as there is no large-scale monolingual corpora for pre-training or sufficient annotated data for fine-tuning. We follow the work of \citet{xia-etal-2021-metaxl} which suggests using meta learning for transfer learning from a single source language to an extremely low resource one. We propose an enhanced approach which uses multiple source languages chosen in a data-driven manner. In addition, we introduce a sample selection strategy for utilizing the languages in training by using a multi armed bandit algorithm. Using both of these improvements we managed to achieve state-of-the-art results on the NER task for the extremely low resource languages while using the same amount of data, making the representations better generalized. Also, due to the method’s ability to use multiple languages it allows the framework to use much larger amounts of data, while still having superior results over the former MetaXL method even with the same amounts of data.

\end{abstract}

\section{Introduction}

Multilingual pre-training such as XLM-R \citet{conneau-etal-2020-unsupervised} have presented great results on various NLP tasks for many languages. But in order to achieve that they require a large scale of monolingual data. Unfortunately, this is not the case for extremely low resource languages such Quechua or Ilocano where data for these languages barely exist. Therefore, in the case of a low resource language a way-to-go approach would be using transfer learning. We follow the work of \citet{xia-etal-2021-metaxl} which uses a meta learning approach. In their work they used a high resource language to learn the task while learning concurrently how to convert representations to the low resource language.

However, the mentioned approach uses only one source language - missing the generalization that could be given using multiple languages. But moving to multi-language approach is not straight forward as different languages can have different effects on the learning process. Selecting languages can be done manually but it is a tedious process that requires linguistic knowledge that sometimes is not widely available. To overcome that, we suggest an approach which uses multiple languages, that are selected in a data driven manner and are balanced during training using a MAB algorithm.

In this paper we show how using multiple languages is more powerful than using a single source language even with the same amount of data. In addition, we propose utilizing multi armed bandit as a sampling strategy to balance the contribution of each language to the training process. Combining both we were able to achieve improved results on the downstream task of NER evaluating languages that are never seen by the pretrained model before. In addition, the language selection is easy and can be done seamlessly.

\begin{table*}[]
\centering
\begin{tabular}{llllllll}
\hline
Source / Target                  & qu             & ilo            & mhr            & mi             & tk             & gn             & Average        \\ \hline
1. English 5k \citep{xia-etal-2021-metaxl}          & 68.67          & 77.57          & 68.16          & 88.56          & 66.99          & 69.37          & 73.22          \\
2. English 20k \citep{xia-etal-2021-metaxl}         & 73.04          & 85.99          & 70.97          & 89.21          & 66.02          & 73.39          & 76.44          \\
3. Related language 5k \citep{xia-etal-2021-metaxl} & 77.06          & 75.93          & 69.33          & 86.46          & \textbf{73.15} & 71.96          & 75.65          \\
4. Related language (6k-8k)               & 76.47          & 82.3           & 73.78          & \textbf{93.53} & 71.07          & 74.07          & 78.54          \\
5. Uniform selection                          & 76.27          & 86.41          & 71.43          & 92.67          & \textbf{72.9}  & \textbf{79.65} & 79.88          \\
6. MetaXLR (ours)                                             & \textbf{78.76} & \textbf{86.96} & \textbf{74.65} & 92.67          & \textbf{73.08} & \textbf{79.44} & \textbf{80.93} \\ \hline
\end{tabular}
\caption{F1 for NER across six settings: 
(1) Source data size of 5k, English data source only.
  (2) Source data size of 20k, English data source only.
  (3) One source language, data size of 5k.
  (4) MetaXL related source language using the exact same data size as we used in our method (varies between 6k-8k as in Table \ref{tab:langs-table}). 
  (5) Choosing languages, uniform distribution. (6) Our method: Source languages defined in Table \ref{tab:langs-table} with MetaXLR algorithm (Algorithm \ref{alg:metaxlr}).}
\label{tab:main-results}
\end{table*}

\section{Method}
\textbf{Using multiple source languages}
To select the source languages we took advantage of both LangRank \citep{lin-etal-2019-choosing} and the languages clusters present in \citet{chiang-etal-2022-breaking}. First, given a target language $t$ we chose a closely related source language $s_1$ used in \citet{xia-etal-2021-metaxl} using LangRank. Next, we used the language clusters and mapped $s_1$ ’s cluster $c$. Then, we chose $n-1$ arbitrary languages from $c$.

\textbf{Multi armed bandit as sampling strategy}
Since the languages from the previous step are selected from a large cluster, they may have different effects on the training process as they are yet varied. Thus, we balance the training process by defining a sampling distribution for the source languages while training.

In our strategy we increase the weight of languages that are harder to learn from. The intuition is that in order to generalize the representations properly across the different languages, we should train more with languages that the model struggles with. For making this weighting strategy adaptive to different languages without manual interference, we reduced the problem to a MAB problem where we consider each source language as an arm. Every training step, we select one language from the language distribution and get a reward, which in this case is the loss - The higher it gets, the more the model struggles with this language. The multi armed bandit we used is EXP3 described in \citet{auer2002nonstochastic}. We used it as part of the meta learning algorithm as can be seen in Algorithm \ref{alg:metaxlr}.

\section{Experiments}

Similarly to \citet{xia-etal-2021-metaxl} we used XLM-R. The data that is used for our experiments is WikiAnn, which contains 282  different languages on the NER task. Results presented in Table \ref{tab:main-results}. Our method outperforms the baselines (1-4) by at least 2.4 F1 score in average, using the same amount of data. Comparing method (1) and (2) emphasizes the importance of the data size. We can also observe the importance of selecting related languages by comparing method (1) and (3). Our method leverages these two observations, by the fact that we are able to use related languages with limited data, and have a large amount of data size combining several related languages together. 
MetaXL \citep{xia-etal-2021-metaxl} works well and improved even by using uniform language selection distribution, by presenting an improved result of at least 1.3 F1 in average. Our language selection algorithm further improves the result by 1.1 F1 score in average. Comparing the two methods (5) and (6), the language selection algorithm performs at least as well as the uniform selection and sometimes outperforms. 

\section{Conclusion}
In this paper, we study cross-lingual transfer learning for extremely low-resource languages. We broadened MetaXL \citep{xia-etal-2021-metaxl}, enabling it to use a set of source languages, while choosing from them in the training loop using a Multi Armed Bandit algorithm. We managed to both improve on the results of previous works while simultaneously increasing the pool of usable data and achieve state-of-the-art results for the extremely low resource languages.

\subsubsection*{URM Statement}
The authors acknowledge that at least one key author of this work meets the URM criteria of ICLR 2023 Tiny Papers Track.

\bibliography{iclr2023_conference_tinypaper}

\begin{thebibliography}{5}
\providecommand{\natexlab}[1]{#1}
\providecommand{\url}[1]{\texttt{#1}}
\expandafter\ifx\csname urlstyle\endcsname\relax
  \providecommand{\doi}[1]{doi: #1}\else
  \providecommand{\doi}{doi: \begingroup \urlstyle{rm}\Url}\fi

\bibitem[Auer et~al.(2002)Auer, Cesa-Bianchi, Freund, and
  Schapire]{auer2002nonstochastic}
Peter Auer, Nicolo Cesa-Bianchi, Yoav Freund, and Robert~E Schapire.
\newblock The nonstochastic multiarmed bandit problem.
\newblock \emph{SIAM journal on computing}, 32\penalty0 (1):\penalty0 48--77,
  2002.

\bibitem[Chiang et~al.(2022)Chiang, Chen, Yeh, and
  Neubig]{chiang-etal-2022-breaking}
Ting-Rui Chiang, Yi-Pei Chen, Yi-Ting Yeh, and Graham Neubig.
\newblock Breaking down multilingual machine translation.
\newblock In \emph{Findings of the Association for Computational Linguistics:
  ACL 2022}, pp.\  2766--2780, Dublin, Ireland, May 2022. Association for
  Computational Linguistics.
\newblock \doi{10.18653/v1/2022.findings-acl.218}.
\newblock URL \url{https://aclanthology.org/2022.findings-acl.218}.

\bibitem[Conneau et~al.(2020)Conneau, Khandelwal, Goyal, Chaudhary, Wenzek,
  Guzm{\'a}n, Grave, Ott, Zettlemoyer, and
  Stoyanov]{conneau-etal-2020-unsupervised}
Alexis Conneau, Kartikay Khandelwal, Naman Goyal, Vishrav Chaudhary, Guillaume
  Wenzek, Francisco Guzm{\'a}n, Edouard Grave, Myle Ott, Luke Zettlemoyer, and
  Veselin Stoyanov.
\newblock Unsupervised cross-lingual representation learning at scale.
\newblock In \emph{Proceedings of the 58th Annual Meeting of the Association
  for Computational Linguistics}, pp.\  8440--8451, Online, July 2020.
  Association for Computational Linguistics.
\newblock \doi{10.18653/v1/2020.acl-main.747}.
\newblock URL \url{https://aclanthology.org/2020.acl-main.747}.

\bibitem[Lin et~al.(2019)Lin, Chen, Lee, Li, Zhang, Xia, Rijhwani, He, Zhang,
  Ma, Anastasopoulos, Littell, and Neubig]{lin-etal-2019-choosing}
Yu-Hsiang Lin, Chian-Yu Chen, Jean Lee, Zirui Li, Yuyan Zhang, Mengzhou Xia,
  Shruti Rijhwani, Junxian He, Zhisong Zhang, Xuezhe Ma, Antonios
  Anastasopoulos, Patrick Littell, and Graham Neubig.
\newblock Choosing transfer languages for cross-lingual learning.
\newblock In \emph{Proceedings of the 57th Annual Meeting of the Association
  for Computational Linguistics}, pp.\  3125--3135, Florence, Italy, July 2019.
  Association for Computational Linguistics.
\newblock \doi{10.18653/v1/P19-1301}.
\newblock URL \url{https://aclanthology.org/P19-1301}.

\bibitem[Xia et~al.(2021)Xia, Zheng, Mukherjee, Shokouhi, Neubig, and
  Awadallah]{xia-etal-2021-metaxl}
Mengzhou Xia, Guoqing Zheng, Subhabrata Mukherjee, Milad Shokouhi, Graham
  Neubig, and Ahmed~Hassan Awadallah.
\newblock {M}eta{XL}: Meta representation transformation for low-resource
  cross-lingual learning.
\newblock In \emph{Proceedings of the 2021 Conference of the North American
  Chapter of the Association for Computational Linguistics: Human Language
  Technologies}, pp.\  499--511, Online, June 2021. Association for
  Computational Linguistics.
\newblock \doi{10.18653/v1/2021.naacl-main.42}.
\newblock URL \url{https://aclanthology.org/2021.naacl-main.42}.

\end{thebibliography}
\bibliographystyle{iclr2023_conference_tinypaper}

\newpage
\appendix
\section{MetaXLR alogrithm}

\begin{algorithm}[h]
\caption{Training procedure for MetaXLR}\label{alg:metaxlr}
\textbf{Input:} Input data from the target language data $D_t$ and the \hl{source related languages $D_{s_1} ... D_{s_K}$}
\\
Initialize base model parameters $\theta$ with pretrained XLM-R weights
\\
Initialize parameters of the representation transformation network $\phi$ randomly
\\
\hl{Initialize weights $w_0(0) = ... = w_0(K) = 1$}
\\
Initialize t = 0
\begin{flushleft}
\begin{algorithmic}[1]
\While {not converged}

\State  \hl{ Define $p_t(i)=(1-\gamma)\frac{w_t(i)}{\sum_{j=1}^{K}}+\frac{\gamma}{K}$}

\State \hl{Sample a source language $s_i$ from} \hspace*{\algorithmicindent} \hl{$p_t(1) ... p_t(K)$}

\State Sample a source batch $(x_{s_i}, y_{s_i})$ from $D_{s_i}$ and \hspace*{\algorithmicindent}a target batch $(x_t, y_t)$ from $D_t$

\State $\theta^{(t+1)} = \theta^{(t)} - \alpha\nabla_{\theta}\mathcal{L}(f(x_{s_i};\theta^{(t)},\phi^{(t)}),y_s)$ 

\State $\phi^{(t+1)} = \phi^{(t)} - \beta\nabla_{\phi} \mathcal{L}(f(x_t;\theta^{(t+1)}),y_t)$ 

\State \hl{$r_t$ =  $\frac{\mathcal{L}(f(x_t;\theta^{(t+1)}),y_t)}{p_t(i)}$}

\State
\hl{$w_{t+1}(i)=w_t(i)e^{\frac{\gamma \cdot r_t}{K}}$}

\State
$t = t+1$
\EndWhile
\end{algorithmic}
\end{flushleft}
\end{algorithm}

The highlighted lines are the main differences between MetaXL and MetaXLR.

\section {Hyper-parameters}
We used $\gamma=0.01$ and 12.5k training steps, with batch size of 4. The rest of the hyper-parameters remained the same as in MetaXL \citep{xia-etal-2021-metaxl}.

\section {Related languages clusters}
\begin{table}[h]
\centering
\begin{tabular}{lll}
\hline
Target Language & Related language in MetaXL & Related languages in our method \\ \hline
qu              & es                         & es, pt, it, de, en, ar, he, fr  \\
il              & id                         & id, he, ar, de, fr, vi, en      \\
mhr             & ru                         & ru, he, ar, de, it, fr, ro, en  \\
mi              & id                         & id, he, ar, de, vi, en          \\
tk              & tr                         & tr, az, be, uk, sk, lt, sr, cs  \\
gn              & es                         & es, pt, it, de, en, ar, he, fr  \\ \hline
\end{tabular}
\caption{Target and source languages information on the NER task. The source data size per source
language is 1k and the target language data size is 100}
\label{tab:langs-table}
\end{table}

\section{Reward strategies}
\begin{table*}[h]
\centering
\begin{tabular}{lll}
\hline
Meta loss as Loss & Uniform Selection & Meta loss as Reward \\ \hline
84.06             & 86.41             & \textbf{86.96}      \\ \hline
\end{tabular}
\caption{F1 score for different reward strategies for the Ilocano target language. Using the loss as a positive or a negative reward affects the performance - rewarding the languages with the high loss, i.e the hard source languages is superior. }
\label{tab:reward-strategies}
\end{table*}

\section{Code}
Our code is available at \url{https://github.com/LiatB282/MetaXLR}

\end{document}